# Genetic Algorithmic Parameter Optimisation of a Recurrent Spiking Neural Network Model

Ifeatu Ezenwe, Alok Joshi, and KongFatt Wong-Lin
Intelligent Systems Research Centre, School of Computing, Engineering and Intelligent Systems,
Ulster University, Magee Campus,
Derry~Londonderry, Northern Ireland, UK
ifeatuezenwe@gmail.com, {a.joshi, k.wong-lin}@ulster.ac.uk

*Abstract*—Neural networks are complex algorithms that loosely model the behaviour of the human brain. They play a significant role in computational neuroscience and artificial intelligence. The next generation of neural network models is based on the spike timing activity of neurons: spiking neural networks (SNNs). However, model parameters in SNNs are difficult to search and optimise. Previous studies using genetic algorithm (GA) optimisation of SNNs were focused mainly on simple, feedforward, or oscillatory networks, but not much work has been done on optimising cortex-like recurrent SNNs. In this work, we investigated the use of GAs to search for optimal parameters in recurrent SNNs to reach targeted neuronal population firing rates, e.g. as in experimental observations. We considered a cortical column based SNN comprising 1000 Izhikevich spiking neurons for computational efficiency and biologically realism. The model parameters explored were the neuronal biased input currents. First, we found for this particular SNN, the optimal parameter values for targeted population averaged firing activities, and the convergence of algorithm by ~100 generations. We then showed that the GA optimal population size was within ~16–20 while the crossover rate that returned the best fitness value was ~0.95. Overall, we have successfully demonstrated the feasibility of implementing GA to optimize model parameters in a recurrent cortical based SNN.

## I. Introduction

To a large extent, information processing in the brain is carried out by dynamic neurons that produce discrete pulses called spikes (i.e. action potentials), exhibiting "all-or-none" responses which influence other neurons connected to them. This led to the development of spiking neural networks (SNNs), treated as the third generation of neural networks [1] [2]. SNNs are thus closer to neurobiological computation than traditional artificial neural networks (ANNs) and have been applied to understanding brain functions and machine learning [1] [2] [3] [4]. From a computational or engineering perspective, SNNs, with their event-driven and spatiotemporal processing, can operate at lower power consumption, faster inference, and massive parallelism than classic ANNs.

Recently, there is highly promising development in applying SNNs to large-scale neuromorphic computing systems [5] [6]. Some of the advantages of neuromorphic hardware are their ability to process data and large amounts of information fast and at lower power. In fact, the realization of SNN in computing neuromorphic systems is seen by many in the field to gain the upper hand for adopting SNNs than traditional ANNs (which still dominate mainstream machine learning).

Despite their exciting developments, there remain several challenges in modelling SNNs. One of them is degeneracy in neural networks. Degeneracy means that elements that are structurally dissimilar perform the same function. There are three types of degeneracy in neural networks: parametric, dynamical, and structural [7]. Parametric degeneracy is associated with the fact that parameters describing a neural network can take a variety of value combinations that give similar patterns of neural activity in this network. Different network connections may give the same network behaviour or function which results in non-uniqueness in model parameter search. Another related problem in modelling SNNs is the time-consuming process of parameter tuning due to the large number of possible free parameters and their combinations. These difficulties pose great challenges to the understanding, simulation and application of SNNs.

A solution to these problems is the use of optimisation techniques [8]. Some of the popular optimisation techniques include maximum likelihood estimation, gradient descent and genetic algorithm (GA). Among these methods, GA is relatively simple to describe and implement [9]. GA, inspired by Charles Darwin's "Survival of the Fittest" biological evolution theory, is a heuristic, stochastic, randomised search optimisation technique [9] [10]. GA also allows multi-objective optimization, searching through a vast number of different possible solutions to find the best one [9]. GA does this by using parallelism which helps to go through many possible values simultaneously, quickly and efficiently. Consequently, numerous solutions must be assessed on each "generation".

A disadvantage of GA is not knowing whether a global optimal solution is reached, not just reaching a local optimum. However, this issue can be solved with crossover and mutation operators by comparing different mutation rates while also increasing the population of the chromosome [11] [12]. In fact, selection, crossover and mutation are the GA operators. Selection is the process by which certain traits become more prevalent in a species than other traits. Crossover is when two chromosomes break and reconnect but with each other's end piece. And mutation is an alteration in the DNA sequence that makes up a gene. Chromosomes are string representations of the outputs to a problem, and GA starts with a randomly generated population of chromosomes.

Although optimisation using GAs are easy to understand and can be applied to a wide variety of optimization problems, not much work has been done using GAs to optimise the parameters for SNN models with recurrent connectivity structure similar to that found in the cortical part of the brain. Previous work of GA optimization applied to SNNs have, for instance, been focused on simple networks for chemotaxis [13], feedforward networks [14] [15], central pattern generators consisting of recurrent inhibitory networks [16] [17], or homogeneous neurons in gene regulatory evolving SNNs [18]. This work will focus on using GA to optimise parameters in a SNN of a canonical cortical "column" with coupled excitatory and inhibitory spiking neuronal populations [19] [20] [21] [22] to achieve some prescribed population neural firing rate target.

## II. SPIKING NEURONAL NETWORK MODELLING

### A. Spiking neuronal models

There are various spiking neuron models that have been developed and studied with each having their advantages and disadvantages, and their choice also depends on the research questions. Some neuronal models are biologically plausible models but can be complex and costly to simulate while others are simple but lack biological accuracy [23]. These models include Integrate-and-fire (IF) model and its leaky (LIF) type [24] [25], Quadratic IF (QIF) [26], and its generalized form (GLIF) [27]. Lower dimensional models include Morris-Lecar model [28], FitzHugh–Nagumo model [29] [30], Izhikevich model [22], in which the latter is an adaptive QIF, and its exponential version (AdEx) of Izhiekvich's model [31], and spike response model (SRM) [32]. In this work, we shall use the Izhikevich model due to its biologically realistic spiking patterns with low computational cost [23].

### B. Izhikevich neuronal model

The Izhikevich neuronal model is a phenomenological model that readily mimics biologically realistic spiking patterns without involving modelling of a variety of ion channel currents [22]. This is achieved with some neuronal membrane excitable dynamics coupled to dynamics of some recovery variable. Thus, the overall neural dynamics can be described by a set of two coupled differential equations [22] [23]:

$$\frac{dv}{dt} = 0.04\,v^2 + 5\,v + 140 - w + I \qquad (1)$$

$$\frac{dw}{dt} = a(b\,v - w) \qquad (2)$$

where $v$ is the membrane potential, $w$ denotes some recovery variable. Different combinations of the four model parameters $a$, $b$, $c$ and $d$ will determine different dynamics and hence spiking behaviours [22]. In addition, $I$ denotes the linear sum of afferent currents from recurrent connections within the network and external or biased input (e.g. due to stimulus) [33] [34]. We differentiate $I$'s for the excitatory and inhibitory neurons with $I_e$ and $I_i$, respectively. These two current types will be used in our parameter search algorithms.

It should be noted that the Izhikevich model does not have a prescribed hard threshold like the IF neuronal model. When the membrane potential $v$ accelerates as in a spike of activity (i.e. mimicking an action potential), it passes some preset peak value (i.e. peak of an action potential, typically set at value from zero to tens of $mV$) and the membrane potential $v$ is reset to the value $c$ and the recovery variable $w$ to $w + d$. We follow typical values of $c$ of $-65\,mV$ and $d$ of 2 for cortical neurons [22], since a neuron's resting membrane potential typically lies within the interval $-60\,mV$ to $-70\,mV$.

The model parameter $a$ denotes the decay rate for $w$ after $v$ reaches its peak value. A smaller value of $a$ means the recovery will be slower. We followed a typical value of $a$ at 0.02 to mimic cortical neurons. The parameter $b$ describes how sensitive $w$ is to the subthreshold activity of the membrane potential $v$. Larger values increase the coupling between $v$ and $w$, giving possible subthreshold fluctuations and low-threshold spiking dynamics [22]. We followed a typical value of $b$ of 0.2 for cortical neurons [22].

### C. Synapses and network

To implement the connections, or synapses, between the spiking neurons, we assumed fast ionotropic receptor mediated currents. These chemical synapses involve neurotransmitter molecules and they are released into the synaptic cleft when the presynaptic neuronal action potential (spike) arrives [25]. The neurotransmitters then bind to the receptors on the post synaptic neuron. The common neurotransmitters include glutamate and γ-aminobutyric acid (GABA), which generally lead to excitatory and inhibitory currents at postsynaptic (i.e. targeted) neurons [25]. In particular, we assumed fast synaptic dynamics such as those mediated by AMPA receptors (~2-5 ms) and $GABA_A$ receptors (~5-10 ms). Hence, we can approximately assume instantaneous synapses [25]. In other words, in such current-based synaptic model, every presynaptic spike will lead to an instantaneous pulse-like increase or decrease by some value in the postsynaptic current through the term $I$ in Eqn. (1), depending on whether the synapses are effectively excitatory or inhibitory, respectively. Note that for simplicity, we have assumed reliable and not probabilistic synapses. To mimic noise in the network, as in an alive brain, we inject additive Gaussian noise into the system as in [22].

Within the network, we assumed that all neurons are connected to all other neurons, i.e. all-to-all connectivity. We then simulated 1000 spiking neurons, 800 of which were excitatory (e.g. pyramidal) neurons and 200 are inhibitory (e.g. fast-spiking) neurons [22]. MATLAB was used in the model simulation. This was run in an Intel® Core™i5-8250U CPU 1.60 HGz processor with 8 GB RAM and 64-bit operating system. Fig. 1 shows a sample simulated spike rastergram (axes: neuron number vs time) of the network spiking activity simulated over 1000 ms.

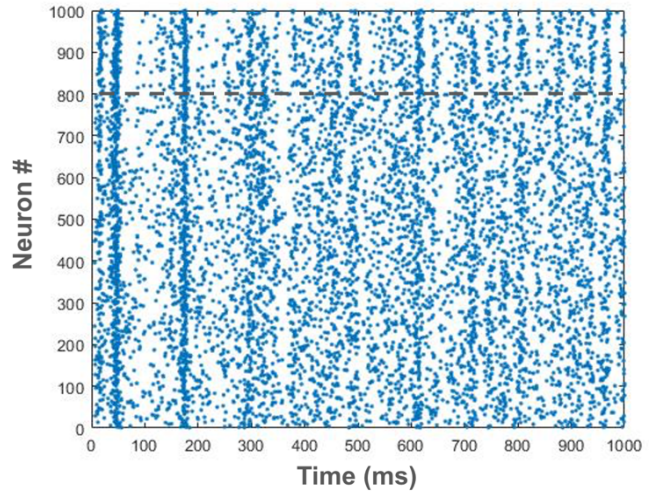

*Fig. 1. Sample simulation of 1000 spiking neuronal network activity over 1000 ms. Vertical (horizontal) axis: neuron number (time in ms). 800 excitatory (bottom 800) neurons and 200 inhibitory (top 200) neurons implemented; neuronal types separated by black horizontal dashed line. Each blue dot denotes a spike of activity for some neuron at some time point.*

## III. GENETIC ALGORITHM OPTIMISATION

We used GA to optimise our SNN model parameters. GA involves population of chromosomes, fitness function, selection proportional to fitness of the chromosomes, crossover and mutation, and have the following steps [9] [10] [11] [12]:

i. Represent every variable as string/chromosome.

ii. State a fitness function to determine how "fit" a chromosome is. The fitness function is used for picking which chromosomes will be selected for mating.

iii. Initialise a population of size N, e.g. $x_1, \ldots x_N$. These are randomly generated be the first generation.

iv. Evaluate the fitness of each chromosome. For maximisation problems, we can use the fitness of the chromosome (string), an objective function $f(x_1, \ldots x_N)$ value. For minimisation problems, we can use the fitness function $\frac{1}{f(x_1, \ldots x_N)}$.

v. Select two chromosomes to mate from the population. Chromosomes are selected to be parents depending on their fitness. In Darwin's theory of natural selection only the fittest natural selection only the fittest survive. So similarly, highly fit chromosomes have a higher probability of being selected for mating than less fit chromosomes. There are various methods used to select which chromosomes will mate. One common one is Roulette wheel selection [36] [37] in which each individual chromosome is given a piece on the wheel. The larger the piece on the wheel the larger the probability of the chromosome being picked. A random number is generated from 0 to100 and the chromosome whose piece is on the number generated will be picked. The first two chromosomes picked will be parents.

vi. The offspring chromosomes are created by using the genetic operators. The crossover operator randomly picks where the two chromosomes should break. It is after that point where the chromosomes will exchange genes. For the mutation operator, GA picks a random gene in the chromosome to invert. So, if it selects a 0 it will invert to a 1 and vice versa. Mutation is done in order to ensure variation in the population in search for better solution. To ensure the algorithm reaches optimal solution, compare results with several different mutation rates (e.g. sometime 0.01 and another one 0.001). To be sure of reliable results, increase the population size as the mutation rates are increased.

vii. Put the produced chromosomes in the updated population.

viii. Substitute the former chromosome population with the new population.

ix. Repeat step 4 until the termination criteria is fulfilled. After various generations, the population reaches a near optimal solution.

These steps are summarised in the flowchart in Fig. 2. We developed MATLAB codes building on MATLAB's GA function, GAOPTIMSET, in the Genetic Algorithm and Direct Search Toolbox [35] (see IV below).

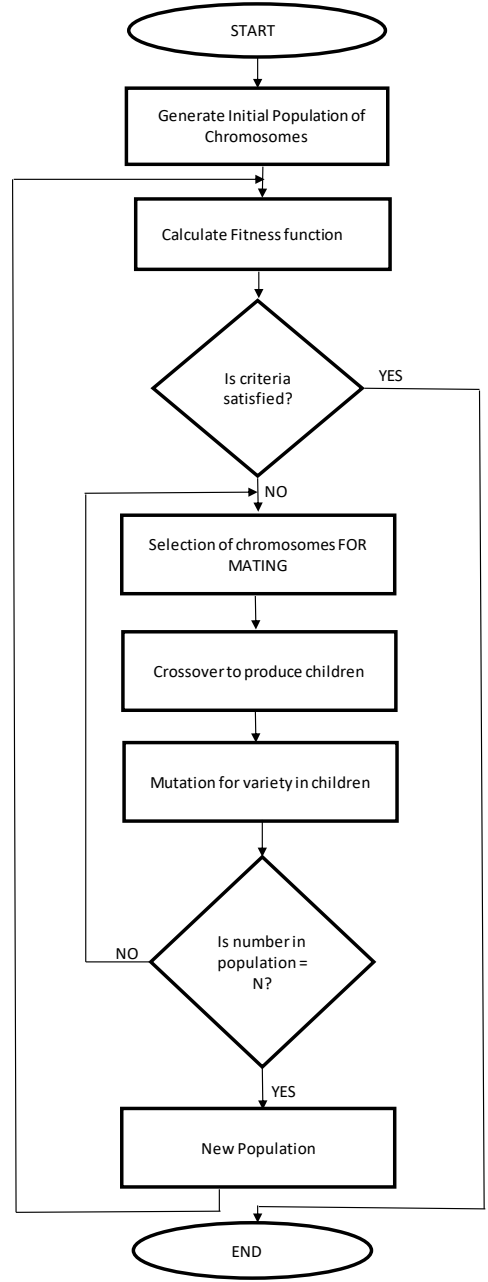

*Fig. 2. Flowchart summarising the steps when implementing GA.*

## IV. GENETIC ALGORITHM ON SNN

Computational neuroscientists or artificial intelligence practitioners often require a certain activity level of the SNN's neuronal population to be set, e.g. to fit to similar values as those observed in wet-lab experiments. Hence, we seek to minimise the error between the simulated and required SNN's population averaged firing rates. We define this error as our function value. In particular, we seek to optimise the biased input current parameter $I_e$ in our SNN with respect to this function value.

Unlike classical ANNs, for SNNs, there is no direct way to apply it to the MATLAB GA toolbox. Modifications were made to the MATLAB code. In particular, the pseudo code for the genetic algorithm applied to SNN to return the value of $I_e$ that will give the smallest difference between the firing rate and the average firing rate is shown below, and the steps illustrated in Fig. 3.

a) Set target firing rate
b) Use GAOPTIMSET function to select desired GA functions
c) Set variable to be optimized
d) Create function handle using difference between simulated average firing rate and target average firing rate calculation
e) Run GA with specified options and function handle

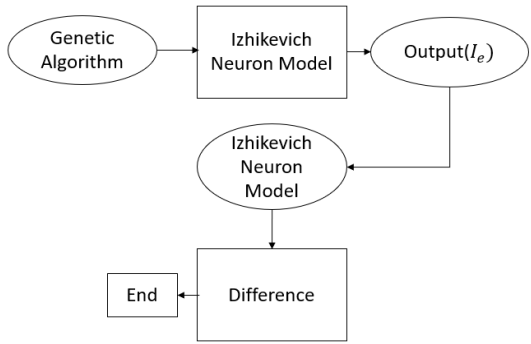

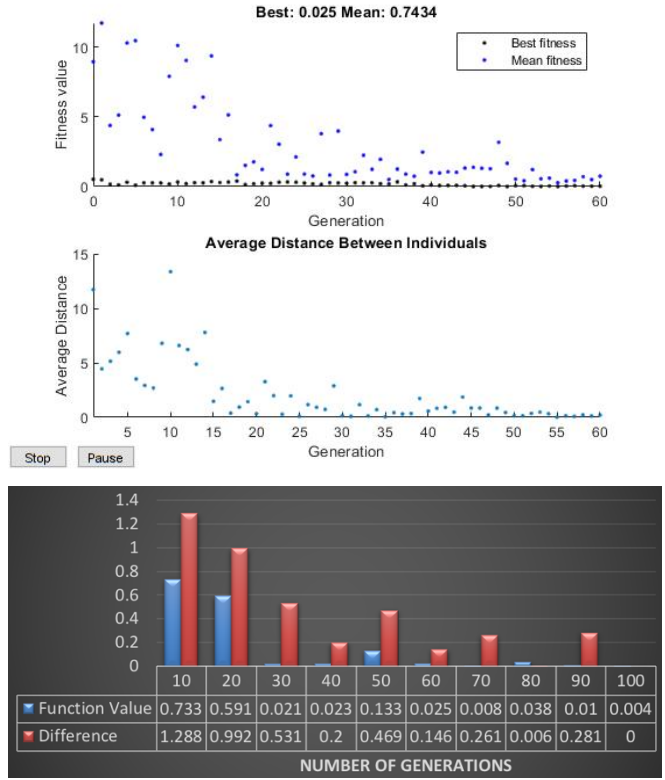

Fig. 3. General steps for applying GA to SNN. SNN comprised Izhikevich neuronal model and current-based synapses. $I_e$ is the biased input current to be searched to minimize the error (GA function value) between the simulated and targeted population averaged firing rates.

Fig. 4. Fitness value (error) and average distance (difference) depend on generation numbers. Top 2 panels: screenshot of a sample fitness value and average distance (difference). Legend in top panel: the best and mean fitness value for each generation. Bottom: aggregated values for different generation numbers.

We set a target population averaged firing rate at 5 spikes/s. The simulated averaged firing rate was calculated by averaging over neurons over a time duration of 1000 ms. We found the optimal $I_e$ values to be ~1.66 (a.u.). (We also successfully did the same for $I_i$ – not shown). After demonstrating the existence of optimal values and focusing on optimizing $I_e$, we then look for gradual convergences of the fitness value (error) and average distance (difference) between individuals over generations. For the latter, the larger the distance between individuals, the higher the diversity. The smaller the distance between the individuals, the lower the diversity. Fig. 4 (top 2 panels) shows a sample of up to 60 generations. The legend in the upper panel shows the best and mean fitness value for each generation while the middle panel shows the average distance between individuals for each generation. Fig. 4 (bottom), which aggregates across several generations, confirms this trend of convergence. Hence, we have demonstrated convergence of the GA on this particular SNN, and a generation number of ~100 minimum is required to obtain the best results.

The GA was then run with crossover fractions from 0 to 1. Crossover fraction is the fraction of the population that crossover will be performed, and mutation is how often genes of the individuals in the population will be mutated. A crossover fraction of 1 means that all the children are a result of the crossover operator while crossover fraction of 0 means that all the children are a result of mutation. The GA was run multiple times with the value of the crossover fraction changed in each run. Fig. 5 (top) shows the effects of crossover fractions with a rate of 0.85. We can clearly observe that as the crossover fraction increased, both the fitness value and average distance (difference) between individuals converged over generations. We found that the crossover fraction that resulted in the lowest function value is ~0.95, before both fitness value and difference shot back up when it reached a value of 1 (Fig. 5, bottom). Note that we could not completely eliminate the fitness value due to intrinsic noise generated in the SNN.

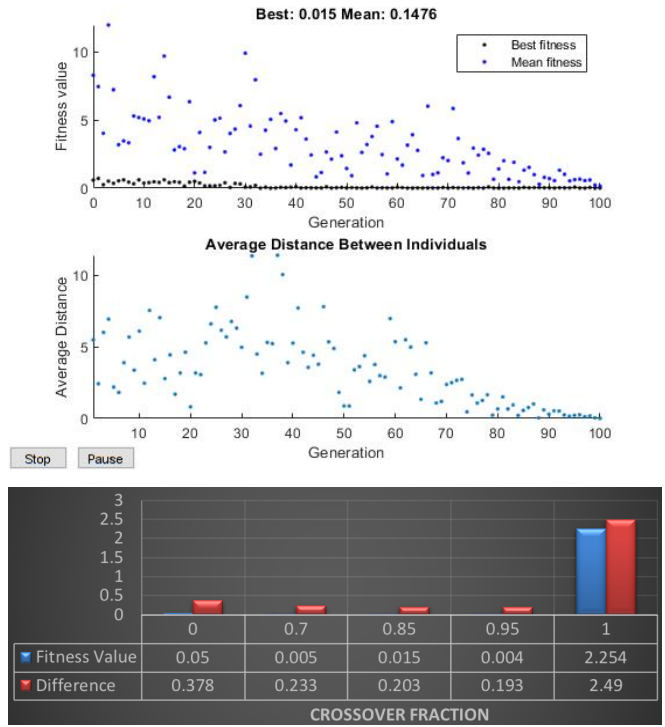

Fig. 5. Fitness value (error) and average distance (difference) vs generation for crossover fraction. Top 2 panels: screenshot of an example with crossover fraction of 0.85. Label as in Fig. 4. Bottom: aggregated values for different crossover fractions.

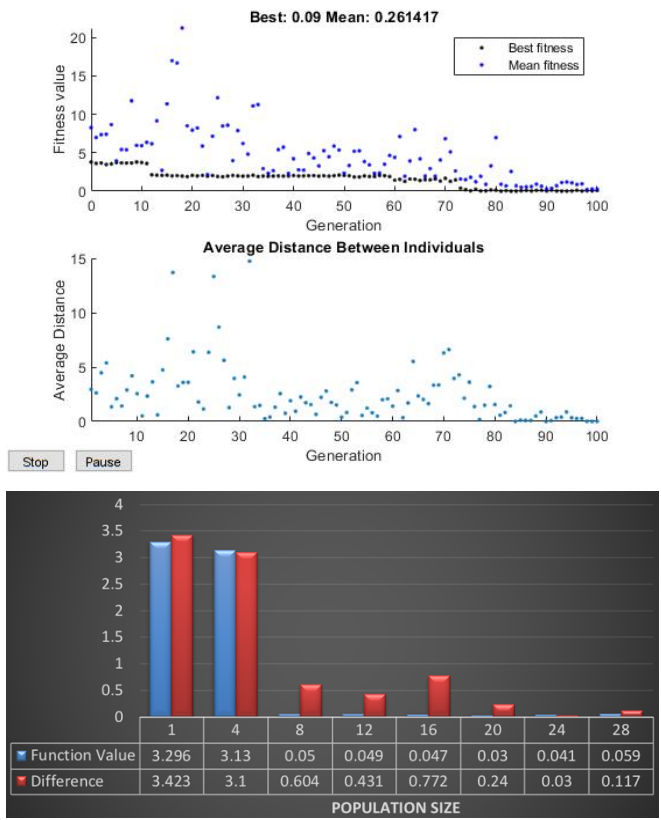

*Fig. 6. Fitness value (error) and average distance (difference) vs generation for population size. Top 2 panels: screenshot of an example with populations size of 12. Label as in Fig. 4. Bottom: aggregated values for different population size numbers.*

We next study the effects of the GA's population size. On the one hand, a higher population size can allow the GA to have a wider search area, and hence higher chance to find the best fitness. On the other hand, higher population sizes take the GA longer to compute each generation and a low population size results in a very high fitness value. For this particular SNN, we found the optimal population size to be within ~16-20 (Fig. 6). Further increase in population size will increase the computational time without substantially improving the fitness values (Fig. 6, bottom).

## V. CONCLUSION AND DISCUSSON

In this work, we have successfully applied GA in search for optimal model parameter of a cortical column-like recurrent SNN consisting of coupled populations of excitatory and inhibitory neurons. In particular, we were able to search for optimal biased input currents to spiking neurons such that the population averaged firing rates were similar to the targeted values. We defined GA fitness function to be the difference between the simulated and targeted firing rates. The optimisation was carried out using modified MATLAB codes for GA and direct search toolbox. Specifically, the algorithm effectively reduced the mean squared error in a converging manner by ~100 generations.

A comparative analysis of the parameters of GA parameters was then performed and analysed in order to find the optimal parameters values which minimised the function value and the averaged distance difference between individuals for the SNN. We showed that the fitness value and averaged distance difference between individuals converged over generation numbers, justifying the approach. Next, we showed that there existed optimal values of crossover fractions (~0.95) and population sizes (~16-20), at least for this particular recurrent SNN.

In this work, we have performed GA based parameter optimisation for only biased input currents. Future work could employ similar methods to search for the synaptic weights of SNNs, or combined synaptic weights and biased input currents (via multi-objective parameter optimization), and not just for simplified firing-rate models (e.g. [39]). In fact, we have successfully performed this for feedforward ANNs (not shown; [38]). Future work should also compare the performances with other methods using the same cortical model, including method using backpropagation through time [40], and explore how such algorithms can be applied to recurrent SNNs in neuromorphic computing system [5].

ACKNOWLEDGMENT

We thank Phil Vance and the rest of the thesis panel for I.E. for constructive feedback on this work.